# MultiProSE: A Multi-label Arabic Dataset for Propaganda, Sentiment, and Emotion Detection

Lubna Al-Henaki[1,2,*], Hend Al-Khalifa[1], Abdulmalik Al-Salman[1], Hajar Alqubayshi[3], Hind Al-Twailay[3], Gheeda Alghamdi[4] and Hawra Aljasim[1]

King Saud University[1], Majmaah University[2], Imam Mohammad Ibn Saud Islamic University[3], Alfaisal University[4]

Riyadh, Saudi Arabia  Lubna.s@mu.edu.sa*

**Abstract**

Propaganda is a form of persuasion that has been used throughout history with the intention goal of influencing people's opinions through rhetorical and psychological persuasion techniques for determined ends. Although Arabic ranked as the fourth most-used language on the internet, resources for propaganda detection in languages other than English, especially Arabic, remain extremely limited. To address this gap, the first Arabic dataset for Multi-label Propaganda, Sentiment, and Emotion (MultiProSE) has been introduced. MultiProSE is an open-source extension of the existing Arabic propaganda dataset, ArPro, with the addition of sentiment and emotion annotations for each text. This dataset comprises 8,000 annotated news articles, which is the largest propaganda dataset to date. For each task, several baselines have been developed using large language models (LLMs), such as GPT-4o-mini, and pre-trained language models (PLMs), including three BERT-based models. The dataset, annotation guidelines, and source code are all publicly released to facilitate future research and development in Arabic language models and contribute to a deeper understanding of how various opinion dimensions interact in news media[1].

## 1   Introduction

Social media has become one of the prevalent mediums for communication channels. Based on recent statistics [1], among 8.08 billion people worldwide, 5.35 billion are internet users, and 5.04 billion are social media users. Therefore, social media platforms have turned into grounds for the rise of proliferation, such as propaganda, which may lead to negative consequences in various domains, including politics, healthcare, and the economy. Based on the Cambridge Dictionary [2], the term propaganda is defined as "information, ideas, opinions, or images, often only giving one part of an argument, that are broadcast, published, or in some other way spread with the intention of influencing people's opinions." Moreover, propaganda utilizes specific rhetorical and psychological persuasion techniques to evoke strong feelings in the audience [3].

Propaganda detection is a crucial task in understanding how information is presented and perceived, particularly in news media. This task involves predicting whether the written text is propagandistic content that is designed to influence public opinion or not. High-performance automated methods for propaganda detection are crucial for enhancing decision-making for authorities, politicians, and businesses. Typically, the main input to propaganda detection models is textual content [4]. However, other inputs, such as sentiment and emotion, can be used to improve the model's performance. It is recommended that more publicly available benchmarked propaganda detection datasets be released under a standard open license. When conducting new propaganda detection datasets, non-English data and annotations of other opinion dimensions, including sentiment and emotions, should be considered.

According to the study's findings in [5], there is a direct correlation between several persuasion techniques and emotional salience features. Furthermore, several studies, such as [6] and [7], [8], [9], and [10], demonstrated the benefits of incorporating sentiment and emotional features in the persuasion detection model. Thus, analyzing the interaction between emotions and propaganda, as well as sentiment and propaganda, could benefit overall propaganda detection. For example, the techniques of "loaded language" and "slogan" exhibit negative correlations with valence and joy intensity while demonstrating positive associations with anger, fear, and sadness intensity [5].

---

[1] https://github.com/xxx/xxx

While efforts on the Arabic propaganda dataset are still in their early stages, the English propaganda dataset has yielded remarkable results. Arabic is the fourth most used language on the internet, with around 237 million Arab users [11]. Moreover, the Arabic language has challenges in NLP because of its complex morphology, orthographic ambiguity, limited availability of linguistic resources (e.g., corpus), and dialectal variances. To the best of our knowledge, no prior work has paid attention to the emotional and sentiment dimensions within the Arabic propaganda dataset.

In this paper, the MultiProSE dataset is released. The first propaganda, sentiment, and emotion annotated corpus specifically for the Arabic language. This dataset comprises 8,000 news texts written in Modern Standard Arabic (MSA) and annotated for propaganda. It is a multi-label extended dataset, with each text also annotated for sentiment and emotion. This comprehensive dataset will serve as a benchmark for three tasks: propaganda detection, sentiment analysis, and emotion recognition. Additionally, it will facilitate the analysis of interactions between various opinion dimensions, including propaganda, sentiment, and emotion. The contributions in this paper can be summarized as follows: (A) We expand the existing propaganda corpus by creating the MultiProSE dataset, specifically designed for the classification of propaganda, sentiment, and emotion in Arabic news texts. This dataset will be made publicly available to support further research in this area. Also, we investigate how propaganda, sentiment, and emotional content are connected and influence each other's impact; (B) We establish robust baselines for the MultiProSE dataset, including three BERT-based models and GPT-4o-mini. The modeling results for each task are reported, providing valuable benchmarks that can guide future research in this domain.

## 2 Related Work

Although propaganda has existed for a very long time, propaganda detection is a relatively new field of study that started in 2017. Several efforts have been conducted to build datasets for propaganda detection tasks. This section will present a comprehensive overview of publicly available datasets, followed by an in-depth analysis of these datasets. Previous studies on propaganda datasets can be categorized based on the level of annotation of the text, which is called granularity. Generally, it is classified into two levels: (1) document level and (2) span level. There are several publicly available datasets for propaganda detection at document level such as TSHP-17 [12], Qprop [13], ProSOUL [14], H-Prop and H-Prop-News [15], and multi-lingual PPN [16].

In addition to document-level datasets, there are also datasets available for propaganda detection at the span level, including Proppy [17], Cazech Propaganda [18], PTC [19], SemEval-2021 Task 6 [20], Arabic Propaganda [21], TWEETSPIN [22], ProText [23], ArAIEval [24], SemEval 2023 Task 3 [25], ArPro [26], X Arabic propaganda [27], PropTweet datasets from Twitter [28], China Propaganda [29]. Table 1 summarizes the publicly available datasets used for propaganda detection. Several conclusions can be drawn from previous studies:

1) propaganda research has expanded beyond English to include other languages, with some multilingual datasets like PPN and SemEval 2023, though most resources remain in English.

2) The two common annotation methods are manual and automated approaches. The quality of manually annotated datasets is higher. It provides a better understanding of the complexities of propaganda, especially in languages with rich morphology, such as Arabic, which is why it is more common. However, manual annotation is time-consuming, relatively small in scale, and expensive. Also, the automated annotation, while it is suitable for large-scale data, is less accurate.

3) most studies using span-level annotation, including sentence or fragment-level, show that this granularity transition from document-level to span-level enhances the understanding of propaganda, allowing for more accurate and detailed identification of propagandistic content.

4) The domain of propaganda datasets includes news articles, social media posts, and memes. Several datasets primarily focus on news articles. While these datasets provide insights into propaganda in traditional media, they cannot address newer forms of digital propaganda. With the increasing influence of social media, several datasets have been constructed from social media platforms.

Table 1: Publicly available datasets for propaganda

| Language | Dataset Name / Ref | Granularity | Dataset Size | Annotation |
|---|---|---|---|---|
| English | TSHP-17 [12] | Document level | 22,580 articles | Weak annotation (distant supervision) |
| | QProp [13] | Document level | 51,294 articles | Weak annotation (Max. entropy) |
| | Proppy [17] | Text span (fragment and sentence) | 350K tokens | Professional annotators |
| | PTC [19] | Text span (fragment and sentence) | 451 articles | Professional annotators |
| | SemEval-2021 Task 6 [20] | Text span (short text) | 950 memes | Professional annotators |
| Arabic | Arabic Propaganda [21] | Text span (short text) | 930 tweets | Professional annotators |
| | ArAIEval [24] | Text span (paragraphs and short text) | 3189 tweets and paragraphs | Professional annotators |
| | ArPro [26] | Text span (fragment and sentence) | 8000 paragraphs | Professional annotators |
| Czech | Czech Propaganda [18] | Both (document and sentence) | 7,494 articles | Professional annotators |
| Urdu | ProSOUL, Humkinar-Web, and Humkinar-News [14] | Document level | 11,574 articles | Weak annotation (ProSOUL is done by Translate QProp into Urdu) |
| Chinese | China Propaganda [29] | Text span (short text) | 9,950 tweets | Professional annotators |
| Hindi | H-Prop and H-Prop-News [15] | Document level | 28,630 and 5500 articles | Weak annotation (Translate QProp into Hindi for H-Prop and use Professional annotators for H-Prop-News) |

5) There is a direct correlation between propaganda techniques and emotional salience features. Analyzing the interaction between emotions and propaganda, as well as sentiment and propaganda, could benefit overall propaganda detection.

In conclusion, the constructed dataset aims to address two significant gaps: language limitations and the incorporation of additional opinion dimensions for more comprehensive propaganda detection. Limited studies have explored the use of opinion dimensions for propaganda detection, particularly in Arabic, which poses unique challenges due to its semantic complexity and the limited availability of NLP resources.

This paper introduced the first Arabic MultiProSE dataset by extending the ArPro propaganda dataset with manually annotated sentiment and emotion labels [26]. This dataset is designed to create a novel Arabic linguistic resource that encompasses propaganda, sentiment, and emotion. Additionally, the MultiProSE dataset employs multi-label classification, which captures multiple aspects of each text. It may significantly enhance its utility for advancing research in this field and enabling the development of more accurate Arabic propaganda detection models.

## 3 MultiProSE Dataset

This section describes the annotation process used to develop the proposed MultiProSE corpus, including the ArPro dataset, the annotation schema and dataset statistics.

### 3.1 ArPro Dataset

The recent Arabic news propaganda benchmark called ArPro was used [26]. This dataset was collected from several Arabic news domains, where part of the dataset was selected from an existing dataset called AraFacts [30] and another part is a large-scale in-house collection. The dataset comprises a total of 8,000 paragraphs, which are divided into training, validation, and testing sets, with 6002, 672, and 1326 samples, respectively. Each sample is associated with a true label if it has at least one propaganda technique; otherwise, a false label is given. Furthermore, 37% of the data is labeled as 'false', and the remaining 63% is annotated as 'true' labels. The dataset encompasses 14 distinct topics including news, politics, health and social. The news and politics topics cover more than 50% of the paragraphs and contains higher amount of propagandistic content.

This dataset was selected because it is the largest Arabic propaganda dataset currently available, offering a rich and extensive collection of data for thorough training and evaluation. Moreover, it exclusively consists of news articles, providing a consistent and pertinent context for analyzing propagandistic content in MSA. The size and specificity of this dataset ensure a comprehensive and accurate approach to propaganda detection.

### 3.2 Annotators information

As recommended by [31] and [32], the MultiProSE corpus was manually annotated by three annotators. All annotators were native Arabic speakers with doctoral degrees: two hold PhDs in Arabic Literature, while the third has a PhD in Criticism and Computational Linguistics. In the emotion annotation task (1- happiness, 2-none, 3-sadness, 4-anger, 5-fear), we use majority voting. If there's a disagreement, adding sixth annotator will help reach a clear decision. Moreover, to guarantee high-quality and accurate results, the annotators were paid[2]. As recommended by [33], the Excel file used for annotation process. To guarantee quality, the annotation guidelines were included in the Excel file.

### 3.3 Annotation Guidelines

The full annotation guidelines are presented in Appendix A. These guidelines are similar to those used in [32], [34], and [35] but have been adapted to suit the specific characteristics of our dataset.

---

[2] The hourly rate is approximately 13.33 USD, and the total time spent annotating all the texts amounts to 80 hours.

Additionally, these guidelines were reviewed by two experts: a PhD holder in Criticism and Rhetoric and an MSc holder in Linguistics, both of whom are native Arabic speakers. Before beginning the annotation process and to ensure the clarity of the guidelines, each annotator participated in a two-hour training session on the annotation guidelines, which were provided in both Arabic and English.

For annotating the dataset, sentiment is annotated with (positive, negative, or neutral) labels similar to [32] and [34]. On the other hand, emotion is annotated based on Paul Ekman's model, which is the most commonly used model in emotion recognition research [36]. It involves six basic emotions: happiness, sadness, anger, fear, surprise, and disgust. However, based on [37] findings, we modified the categories of Ekman's model to include the "none" label, indicating that the text does not express any emotion. We omitted the disgust category due to its confusion with anger. Therefore, the final emotion labels are (happiness, sadness, anger, fear, or none). As mentioned by the authors [38], the labeling of sentiment is correlated with emotion labels. This means that a text annotated as positive will have happiness as the emotion label, while a text with negative sentiment will be annotated with anger, sadness, or fear as emotion labels. Therefore, in the annotation guidelines, these details have been considered and incorporated to ensure that sentiment is annotated independently of the emotions in the text and vice versa. Also, to avoid any influence on the annotators' decisions, each text is annotated separately for emotion and sentiment at different times.

### 3.4 Quality Control Mechanism

To guarantee the quality of the annotations, the quality control mechanism (QCM) based on inspiration from [39] is followed. The main aim of QCM is to ensure accuracy and reliability by preventing random text annotation. The following points describe the four steps of the QCM process.

- **Gold Data Phase:** Similar to the criteria used in [35], a subset of 5% of the corpus, about 400 texts, is randomly selected as ground truth for QCM purposes. This ground truth is annotated by one of the authors (annotation procedure manager), and to ensure quality, it is also reviewed by two Arabic language experts—one with a PhD in Arabic Literature and Criticism, and the other with a master's degree in linguistics. During the annotation phases, if the annotators annotated these gold data incorrectly, they were informed of the mistake. However, if the annotator's trust score, which measures an annotator's consistency and reliability to the provided guidelines, is below 70%, all their submitted annotations are eliminated and ignored.

- **Training Phase (Pre-Exam):** A pre-examination was given to 15 candidate annotators to assess their understanding and skill. Before the pre-exam, several informal meetings were conducted to explain the annotation guidelines, give some examples of annotation, and then discuss the possible challenges. From the gold data, a subset of 100 texts is selected at random to serve as quality test data. After the pre-exam, annotators are qualified to participate in annotation phases if they achieve an accuracy exceeding 70% in both sentiment and emotion annotation tasks [35]. Finally, the three annotators are selected based on the highest scores achieved.

- **During Annotation Tasks:** In actual annotation tasks, in each annotation round, 100 texts are randomly selected from ground truth and injected in all annotated text. After that, the weak annotation scores below 70% are discarded. Although this quality control added an extra effort and increased the annotation period, it provides greater reliability of the annotated data quality.

- **After Annotation Task** After each phase of annotation, the annotation manager calculates the quality of annotations by computing the Inter-Annotator Agreement (IAA), as will be explained in the next section. Next, a deep analysis is performed on these results to identify any conflicts and challenges in the annotations.

## 3.5 Annotation Phases

This section provides details of the annotation phases. Figure 1 present the annotation process pipeline. The annotators started the corpus annotation on June 10, 2024, and completed it by September 14, 2024. The corpus was divided into several rounds, with each round containing several batches. Each round includes approximately 2,600 texts, divided into 4 batches. Each text was independently annotated by three annotators. At the end of each batch, an evaluation of the annotation quality was conducted to ensure that only high-quality annotators, as described in the previous section, were used. By the end of the 12 batches, annotations for 24,000 texts, with a ratio of 3 annotations per text, were collected.

- **Pilot Run:** The dataset was extracted from Arabic news domains, as described in Section 3.1, which requires annotators to have a collective understanding of what constitutes sentiment and emotion. Therefore, the first phase is called the pilot round, which involves a smaller subset of the corpus, around 600 texts, to ensure that the annotation guidelines are clear and understood by all annotators. After analyzing the results of this phase, some minor differences among the annotators were detected, due to the subjectivity of the task itself. Consequently, the annotation guidelines were revised to address all discrepancies.

- **Revised Instructions:** Each annotator annotated the remaining texts to ensure that the updated annotation guidelines were explicit and detailed. In this stage, the annotators annotated the texts based on the updated set of guidelines annotation. The results of this phase helped in the revision of the guidelines to cover many styles of news text and guaranteed the annotators' ability to consistently recognize sentiment and emotions.

- **Consolidation Phase:** After each round in the previous section, the annotation manager analyzed the disagreements among most of the annotators and gathered feedback to maximize quality and consistency. Additionally, a majority voting scheme was used to determine the final emotion and sentiment labels for each text. As mentioned earlier, each text is annotated by three annotators; in case of disagreement, up to six people may be involved in making the final decision.

## 3.6 Inter-Annotator Agreement

After annotation, the reliability of the annotation scheme was assessed using IAA. Cohen's Kappa is among the most widely used metrics for assessing agreement between two annotators on categorical variables [40]. As we have more than two annotators, as described in the annotator's information section, Light's Kappa is particularly suited for three or more annotators. It involves calculating Cohen's Kappa for every possible pair of annotators involved in the annotation task and then averaging these values to obtain the overall agreement among all annotators [41]. Moreover, Fleiss' Kappa [42], another extension of Cohen's Kappa specifically used to measure agreement between more than two annotators, is used. The authors in [43] and [44] indicate that human annotators agreed approximately 70%-80% on binary or ternary classes. However, as the number of classes increases, it becomes more challenging for annotators to reach agreement. The MultiProSE corpus consists of 3 and 5 classes for sentiment and emotion, respectively. Therefore, this highlights that multi-label sentiment and emotion annotation is a complex task for humans to agree on. The IAA for emotion annotations, as indicated by Lights' index and Fleiss' Kappa, are 0.7074 and 0.7093, respectively. In contrast, sentiment analysis demonstrates higher agreement with values of 0.8128 for Lights' index and 0.7650 for Fleiss' Kappa. According to [40], this value is interpreted as substantial.

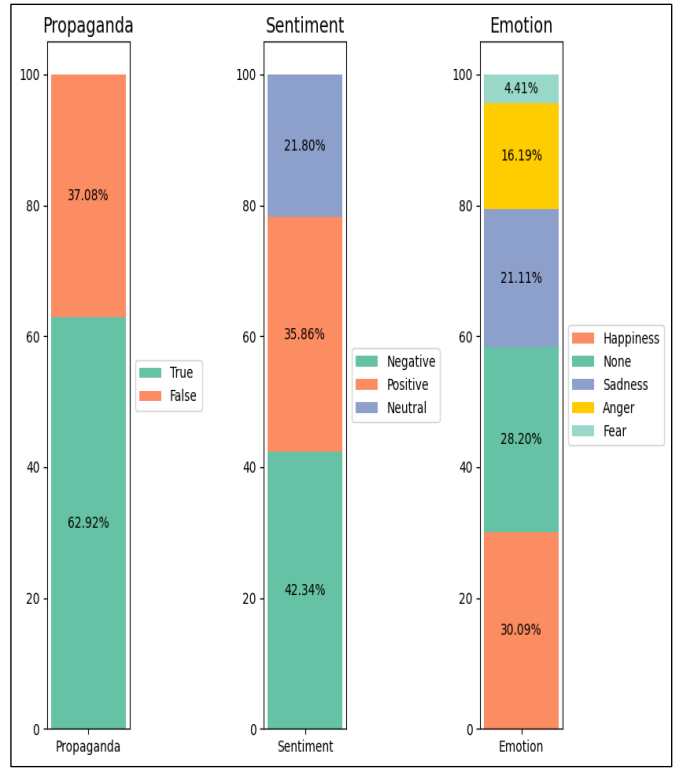

Figure 2 Labels' Distribution Across All Tasks

### 3.7 Dataset Statistics

The MultiProSE is a multi-label dataset containing 8000 annotated texts, where each text is annotated for propaganda, sentiment, and emotion. Table 2 shows some examples from the MultiProSE dataset, particularly highlighting cases where annotated sentiment is not influenced by emotion. The dataset is split into 83.5% (6680 texts) and 16.5% (1320 texts) for training and testing, respectively, as presented in Table 3.

Figure 2 illustrates the labels' distribution across all tasks. As observed in Figure 2, the percentage of texts that have a fear emotion is low, 4.41%, compared to those labeled as happiness 30.09%. This can be explained by the fact that happiness includes various emotions, such as joy, cheerfulness, satisfaction, contentment, and fulfillment. Moreover, negative emotions, such as sadness and anger, have a significant presence with 21.11% and 16.19%, respectively. Furthermore, negative sentiment is the most frequent, comprising over 42.34% of the data. Positive sentiment follows closely at 35.86%, indicating a relatively balanced distribution of emotional polarities. Texts marked as propaganda dominate the dataset, constituting nearly two-thirds of all instances.

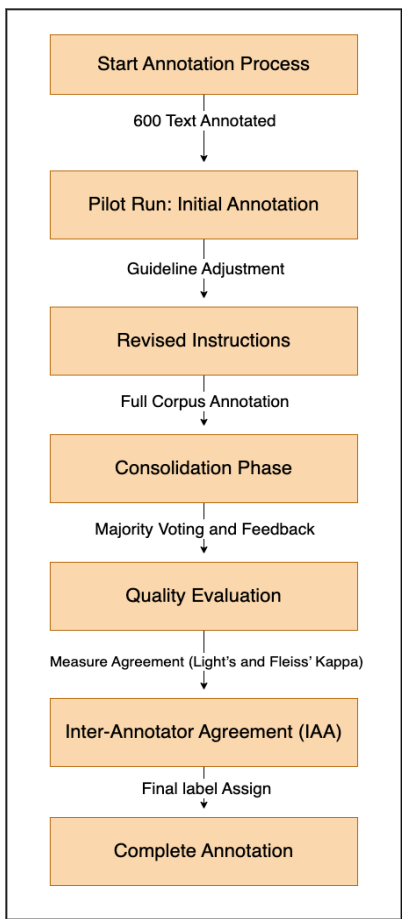

Figure 1 Annotation Process Pipeline

Table 2: Example from MultiProSE dataset translated into English

| Text | | Propaganda | Sentiment | Emotion |
|---|---|---|---|---|
| Sheikh Zayed decided to cut off oil to Israel, saying: "Oil is not more precious than Arab blood," and this was a strong pressure factor on foreign countries. | قرر الشيخ زايد قطع النفط عن إسرائيل قائلاً: «النفط ليس أغلى من الدم العربي»، وكان هذا عامل ضغط قوي على الدول الأجنبية. | True | Positive | Anger |
| Schedule of departure and arrival times of international and domestic flights to and from Egypt International Airport on Monday | جدول مواعيد الإقلاع والوصول للرحلات الدولية والداخلية من وإلى مطار مصر الدولي اليوم الاثنين | False | Neutral | None |
| He stressed that the Corona pandemic crisis will create more turmoil at the global level. | وأكد أن أزمة جائحة كورونا ستخلق مزيداً من الاضطرابات على المستوى العالمي. | True | Negative | Sadness |

Table 3: Data split statistic of MultiProSE dataset

| Task | Label | Train | Test |
|---|---|---|---|
| Propaganda | True | 62.99% | 62.9% |
| | False | 37.01% | 37.1% |
| Sentiment | Positive | 35.60% | 37.4% |
| | Negative | 42.70% | 40.4% |
| | Neutral | 21.70% | 22.2% |
| Emotion | Happiness | 29.80% | 31.5% |
| | Sadness | 20.40% | 24.5% |
| | Anger | 16.50% | 14.4% |
| | Fear | 4.60% | 3.4% |
| | None | 28.60% | 26.2% |

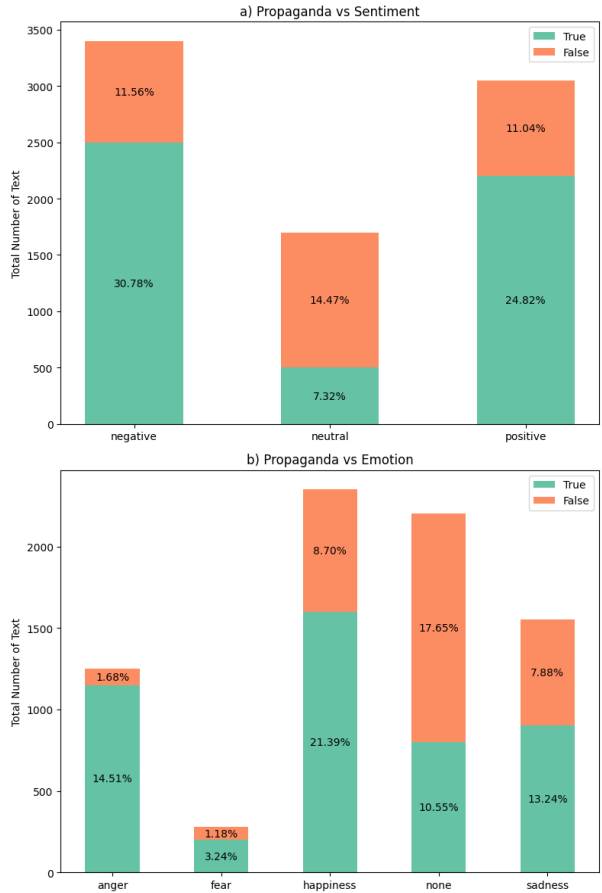

Figure 3: Distribution of (a) Sentiment and (b) Emotion Labels by Propaganda

Figure 3 presents the distribution of sentiment and emotion per propaganda. This finding confirms the findings of several studies, including [5], [6], [7], [8], [9], and [10]. These studies demonstrate a direct correlation between propaganda and emotional salience features, highlighting the benefits of incorporating sentiment and emotional features in propaganda detection models. Furthermore, Figure 3 clearly shows that propaganda is more frequent in texts with negative and positive sentiment, particularly in negative sentiment at 30.78% and positive sentiment at 24.82%, compared to only 7.32% in neutral sentiment. Additionally, propaganda is most associated with anger and happiness, at 14.51% and 21.39%, respectively. In contrast, fear and sadness contain fewer propaganda paragraphs. Furthermore, texts without propaganda exhibit a higher proportion of neutral sentiment and no emotion, with 14.47% and 17.65%, respectively, compared to texts with propaganda, which show 7.32% for neutral sentiment and 10.55% for no emotion.

## 4 Benchmark Experiments

This section presents a strong baseline model in propaganda, sentiment and emotion tasks intended to motivate and support researchers in the development of advanced models for text analysis across these domains. A MultiProSE dataset is suitable for classification tasks only when the accuracy of its annotations is verified. To achieve this, this section details the setup and design of experiments.

### 4.1 Models

Pre-trained language models (PLMs) and large language models (LLMs) are developed. For LLMs, the GPT-4o-mini is finetuned [45]. For PLMs, Bidirectional Encoder Representations from Transformers (BERT) based models have shown significant performance in diverse NLP classification tasks [46]. Thus, several BERT-based classifiers for propaganda, sentiment and emotion tasks are developed. Specifically, two Arabic state-of-the-art transfer learning models are finetuned to evaluate the annotations as follows:

- **AraBERT** is trained on 8.6 billion tokens from five datasets consisting of MSA text [47].

- **XLM-RoBERTa** is a multi-lingual language model trained on 2.5 terabytes of text across 100 languages, including Arabic [48].

Similar to [26], the dataset is split into 75%, 8.5%, and 16.5% for training, development, and testing, respectively. The transformer toolkits are employed to develop a pipeline for propaganda, sentiment, and emotion analysis [49]. The finetuning code is available online, accompanied by the MultiProSE dataset.

For PLMs, the proposed system initiates preprocessing Arabic texts by applying character normalization, removing diacritics, stopping words, tatweel, non-Arabic letters, and repeating characters. Based on the training data visualization, the maximum sequence length is set to 256 tokens. Additionally, after conducting multiple empirical experiments, the batch size and number of epochs were set to 8 and 5, respectively. The AdamW optimizer was used with a learning rate of 2e-5 [50]. For LLM, the temperature value was set to zero to get deterministic and accurate decision-making. Also, the batch size and number of epochs were set to 8 and 4, respectively. Also, for each experiment, six and three runs with different random seeds were conducted for PLMs and LLM, respectively, and then the average performance over the test subset was reported.

### 4.2 Results and Discussion

Table 4 presents the obtained results of the proposed models, highlighting the best results for each task in bold text. It is clear from the table that AraBERT outperforms the other models in propaganda detection, achieving the best overall performance across most evaluation metrics. Nevertheless both AraBERT and GPT-4o-Mini obtaining a Micro-F1 score of 0.769. However, XLM-RoBERTa performed the worst in this task, yielding a Micro-F1 score of 0.683.

For sentiment analysis, GPT-4o-Mini achieved the highest score with a Micro-F1 score of 0.842, followed by AraBERT with a slight difference between them in Micro-F1 scores by 0.002. Again, XLM-RoBERTa achieved a lower performance with a 0.698 Micro-F1 score. On the other hand, in emotion detection, GPT-4o-Mini demonstrated strong performance compared to the remaining models, with a Micro-F1 score of 0.750. AraBERT followed with similar performances, yielding Micro-F1 scores of 0.675.

Overall, the GPT-4o-Mini outperforms the other models across most tasks, especially sentiment analysis. Also, ArabicBERT achieved results comparable to those of AraBERT. This highlights the potential of the type of training data (i.e., dialectal and MSA), which reflects MultiProSE properties, as it contains a small portion of dialectal news. Furthermore, AraBERT exceeds it in some tasks due to the size of the trained model. However, XLM-RoBERTa is the worst model across all tasks. This may be explained by the fact that XLM-RoBERTa is multilingual in nature and not trained specifically in Arabic. In terms of overall task complexity, the emotion detection task seems to be more challenging for all models due to complex linguistic patterns and the number of emotions involved.

Table 4: MultiProSE results on test set. Acc is an abbreviation for Accuracy.

| Task | AraBERT | | | XLM-RoBERTa | | | GPT-4o-Mini | | |
|---|---|---|---|---|---|---|---|---|---|
| | Micro-F1 | Macro-F1 | Acc % | Micro-F1 | Macro-F1 | Acc % | Micro-F1 | Macro-F1 | Acc % |
| Propaganda Detection | **0.769** | **0.756** | **77** | 0.683 | 0.597 | 68 | 0.769 | 0.733 | 76 |
| Sentiment Analysis | 0.736 | 0.722 | 73 | 0.698 | 0.682 | 69 | **0.842** | **0.825** | **84** |
| Emotion Detection | 0.675 | 0.635 | 67 | 0.648 | 0.608 | 64 | **0.750** | **0.707** | **75** |

## 4.3 Conclusion

In this paper, MultiProSE, is presented, as the first multi-label Arabic dataset for propaganda detection. The dataset is an extension of the ArPro dataset, with each text annotated for sentiment and emotion. Thus, MultiProSE can serve as a new benchmark for three tasks: propaganda detection, sentiment analysis, and emotion recognition. Moreover, it can enable future research to study the interaction between different opinion dimensions. A detailed description of the annotated dataset and a statistical analysis of the produced annotations are presented. Finally, several experiments for each task have been developed based on GPT-4o-mini and two BERT-based models. For future work, annotation based on span-level analysis and building a lexicon may provide deeper insights and boost the performance of detection models.

## Limitations

During the annotation process, several challenges were faced by the annotators, such as:

- **Time Constraints:** while the expected annotation time for each text is calculated during the gold data phase, there was some overestimation of the annotators' ability to manage such a large corpus. The annotation tasks were completed within four months. Following the pilot study, annotators encountered difficulties balancing accurate annotation with the deadlines. To address this challenge, each annotator was assigned a minimum daily goal. The minimum number of texts per day was set to 100 texts, which encouraged annotators to maintain consistency.

- **Annotation Guidelines:** As the corpus was extracted from news articles from a variety of Arabic news domains, developing clear and detailed guidelines that cover all possible scenarios in the corpus was a significant challenge.

- **Diverse Topics Covered:** The ArPro corpus covers a wide variety of 14 different topics, such as news, politics, health, and sports. Therefore, annotating emotions and sentiments across different topics requires an in-depth understanding of the context and the emotions expressed, especially for news and politics, as they comprise more than 50% of the corpus.

- **Diacritical Marks:** The lack of diacritical marks in most of the paragraphs leads to some words being ambiguous and having multiple possible meanings, which can result in inaccurate annotation. This challenge was addressed by asking the annotators to understand the term in relation to its context or refer back to the source of the news for a comprehensive understanding.

- **MSA and Dialect:** Most of the paragraphs are written in MSA; however, a few are presented in dialect because they include quotes from individuals in the news. Consequently, annotators encountered difficulties with some obscure dialectal words that are rare in certain regions, necessitating additional searches for their meanings. Furthermore, some MSA words were ambiguous and unclear.

- **Typos and Grammar:** A few paragraphs have spelling and grammatical mistakes, which

create ambiguity in the text by changing its meaning

# Acknowledgments

This research was accomplished by utilizing the technical and linguistic resources at the Thakaa AI Center for Arabic and King Salman Global Center for the Arabic Language in Riyadh, Saudi Arabia.